\pdfoutput=1
\documentclass[11pt]{article}

\usepackage{emnlp2021}

\usepackage{times}
\usepackage{latexsym}

\usepackage[T1]{fontenc}
\usepackage[utf8]{inputenc}

\usepackage{microtype}

\usepackage{hyperref}       % hyperlinks
\usepackage{url}            % simple URL typesetting
\usepackage{booktabs}       % professional-quality tables
\usepackage{amsfonts}       % blackboard math symbols
\usepackage{nicefrac}       % compact symbols for 1/2, etc.
\usepackage{microtype}      % microtypography
\usepackage{comment}
\usepackage{natbib}
\usepackage{multirow}
\usepackage{subfigure}
\usepackage{subfiles}
\usepackage{graphicx}
\usepackage{todonotes}
\usepackage{algorithm2e}
\usepackage{mathabx}

\usepackage{float}
\title{Multi-Sentence Resampling: A Simple Approach to Alleviate Dataset Length Bias and Beam-Search Degradation}

\author{Ivan Provilkov \\
  Yandex Research, Moscow \\
  Moscow Institute of Physics and Technology \\
  \texttt{iv-provilkov@yandex-team.ru} \\\And
  Andrey Malinin \\
  Yandex Research, Moscow \\
  HSE University, Moscow \\
  \texttt{am969@yandex-team.ru} \\}

\begin{document}
\maketitle
\begin{abstract}
% Neural Machine Translation is known to suffer from a beam-search problem: more probable hypotheses obtained with growing beam sizes have low quality. Higher likelihoods are given to incorrect short translations ignoring correct hypotheses. This property drops quality, especially for long sentences causing the length bias. While much work is done on these problems, there is no answer to why models provide such probabilities biased towards short incorrect translations. In this work, we analyze errors that cause major quality degradation with large beams and show that the reason for the beam-search problem and length bias \textit{is Machine Translation datasets that are significantly biased towards short sentences}. To mitigate this issue, we propose a new data augmentation technique -- \textit{Multi-Sentence Resampling}. This technique modifies original training examples by concatenating several sentences from a dataset to make a long training example. Using \textit{Multi-Sentence Resampling} significantly reduces beam-search problem and length bias and improves final translation quality.

Neural Machine Translation (NMT) is known to suffer from a beam-search problem: after a certain point, increasing beam size causes an overall drop in translation quality. This effect is especially pronounced for long sentences. While much work was done analyzing this phenomenon, primarily for autoregressive NMT models, there is still no consensus on its underlying cause. In this work, we analyze errors that cause major quality degradation with large beams in NMT and Automatic Speech Recognition (ASR). We show that a factor that strongly contributes to the quality degradation with large beams is \textit{dataset length-bias} - \textit{NMT datasets are strongly biased towards short sentences}. To mitigate this issue, we propose a new data augmentation technique -- \textit{Multi-Sentence Resampling (MSR)}. This technique extends the training examples by concatenating several sentences from the original dataset to make a long training example. We demonstrate that MSR significantly reduces degradation with growing beam size and improves final translation quality on the IWSTL$15$ En-Vi, IWSTL$17$ En-Fr, and WMT$14$ En-De datasets. 
\end{abstract}

%~\citep{cat-tongue}~\citep{correcting-length-bias-2018}

\section{Introduction}
%\todo{Introduction to problem, its importance, current solutions: empirical methods, some interesting analysis, our contribution}

% What is the source?
% Previous works performed big analysis. They show that the beam-search problem is closely connected with too high probabilities that the model gives to short sentences and how bad probabilities can cause search errors and quality degradation with growing beam size. However, there is still no answer for one of the main questions: \textit{Why do NMT models produce bad probabilities?}

%Intro to problems

In this work, we address the \textit{beam-search problem} in Neural Machine Translation~\citep{mt-problems-2017}. Beam Search is the standard hypothesis search method for autoregressive sequence generation. Large beams provide more probable hypotheses than small beams; however, the overall translation quality drops with growing beam size after a certain point. This effect is especially strong for long sentences, connecting with the fact that NMT models are biased to giving high probabilities to short hypotheses. ~\citet{cat-tongue} showed that exact search by likelihood for neural machine translation finds empty string as the optimal hypothesis in more than $50 \%$ of cases.

%In this work, we address several Neural Machine Translation problems~\citep{mt-problems-2017}. The first problem is called a \textit{beam-search problem}. It is connected with another widely-known problem called \textit{length bias}. That is that most likely hypotheses of a translation model tend to be short and meaningless.~\citep{cat-tongue} showed that exact search by likelihood for neural machine translation finds empty string as the optimal hypothesis in more than $50 \%$ of cases. This causes a dramatic drop in translation quality for long sentences~\citep{mt-problems-2017}. 

% Empirical methods
One of the most famous methods to mitigate quality degradation with growing beam size is length normalization~\citep{bahdanau2016neural, Wu2016GooglesNM}. This technique normalizes log-likelihoods of a hypothesis in beam search by its length, thus promoting long hypotheses. Other methods examine adding a reward for each token's score during the decoding process~\citep{breaking-beam-curse-2019, correcting-length-bias-2018}.

%While these works proposed empirical methods to decrease the effect of length bias and beam-search problem, they do not provide an explanation of reasons of these problems. 

%Overestimated prefixes and analysis
While the beam-search problem has been extensively studied~\citep{2016-length-bias, correcting-length-bias-2018, calibration-of-encoder-decoder, beam-search-question, map-decoding-all-you-need, predicting-length-2020, wang-sennrich-exposure-2020} there is still no consensus on the underlying reason for such model behavior. Furthermore, prior work has investigated this problem primarily for NMT models, giving little attention to other domains that are also known to suffer from it, such as Automatic Speech Recognition (ASR)~\citep{towards-better-decoding-asr, renormalize-prob-pruning-beam-asr}. \citet{correcting-length-bias-2018} noticed that since in each step of beam-search generation, negative log-probability is added to the hypothesis' score, if a model overestimates the probability of an already generated sequence of tokens, there is no way to downgrade this probability afterward. Consequently, models are biased towards finalizing a short hypothesis by generating an end-of-sequence token (EOS) rather than generating a long continuation. Their experiments show a connection between quality degradation and decreasing length of hypotheses with growing beam size. Our work is in agreement with their explanation. Moreover, we show that the main quality degradation with large beams in NMT and ASR comes from short translations obtained by early termination of long hypotheses from small beams.

% The source is datasets
Our work examines how the distribution of sentence lengths in a dataset affects the beam-search problem. We demonstrate that the \textit{ beam-search problem is strongly connected with the distribution of sentence lengths in training datasets}. Specifically, we show that common NMT datasets, such as IWSLT and WMT, exhibit a strongly skewed distribution of sentence lengths, with the mode focused on short sentences. NMT models learn biased probability distributions and fail on long sentences during inference. In contrast, for ASR models trained on Librispeech~\cite{librispeech}, where the distribution of sentence lengths is more symmetrical and biased towards longer sentences, the beam-search degradation occurs at much larger lengths. Based on our findings, we propose a simple and effective dataset augmentation technique that makes training examples longer -- \textit{Multi-Sentence Resampling}. It creates a new dataset where each training sample can be a concatenation of multiple sentences. Our method alleviates quality degradation with growing beam size and increases the final quality of the model. 

%Key contributions
The key contributions of our work are as follows:
\begin{itemize}
    \item We show that quality degradation with growing beam size comes mostly from short translations, which are early finalized prefixes of long hypotheses;
    \item We show that training datasets that are biased towards short sentences strongly contribute to the beam-search problem;
    \item We introduce \textit{Multi-Sentence Resampling} -- a simple and effective dataset augmentation technique that alleviates beam search problem and increases the final translation quality\footnote{Our code is available at \url{https://github.com/yandex-research/msr}}.
\end{itemize}

\section{Quality degradation analysis}

\begin{table*}[t!]
\centering
\small
\begin{tabular}{lccccc}
\toprule
 & & Number of sentences& Number of tokens & Avg. sentence length &  \\
 & & (train/dev/test)& in train (en) & in tokens &  \\

\midrule
%\multicolumn{5}{l}{\bf IWSLT15}\\
%IWSLT15 & & & &\\
%IWSLT15& En $\leftrightarrow$ Vi & 133k / 1553 / 1268 & 4k & 4k & 0.1 / 0.1\\
%& En $\leftrightarrow$ Zh & 209k / 887 / 1261 &  4k / 16k  & 4k & 0.1 / 0.6\\
%\cmidrule{2-6}
%\multicolumn{5}{l}{\bf IWSLT17}\\
%IWSLT17 & & & &\\
Librispeech& clean & 133k / 2703 / 2620& 4{.}6M & 34{.}9 & \\
% & other & 149k / 2864/ 2939& 4{.}8M & 32{.}4 &  \\
\cmidrule{2-6}
IWSLT17& En $\leftrightarrow$ Fr & 232k / 890 / 1210& 4{.}8M  & 20{.}5 &\\
%& En $\leftrightarrow$ Ar & 231k / 888 / 1205 & 4k & 4k & 0.1 / 0.1\\
\cmidrule{2-6}
IWSLT15& En $\leftrightarrow$ Vi & 133k / 1553 / 1268 & 2{.}7M & 20{.}3 & \\
\cmidrule{2-6}
%\multicolumn{5}{l}{\bf WMT14}\\
WMT14 & En $\leftrightarrow$ De & 4{.}5M / 3000 / 3003& 11{.}4M  & 25{.}2 &\\

%& En $\leftrightarrow$ Fr & ?? & ?? & ?? \\
%\cmidrule{2-6}
%\multicolumn{5}{l}{\bf ASPEC}\\
%ASPEC & En $\leftrightarrow$ Ja & 2M / 1700 / 1812 & 16k & 32k & 0.1 / 0.6\\
\bottomrule
\end{tabular}
\caption{Overview of the datasets. The number of tokens is calculated after Moses preprocessing\footnote{\url{https://github.com/moses-smt/mosesdecoder}}.}
\label{tab:datasets}
\end{table*}

%\todo{Here we compare NMT vs ASR.Firstly show that quality degradation comes from prefixes, then that quality degradation correlates with number of train examples, and that beam-search degradation is likely to be due to length distribution of train datasets.}

%While a lot of work done in the field of length bias and beam search in Neural Machine Translation, and many heuristics proposed to overcome these issues, it is still not clear what is the reason for these problems, and how exactly drop in quality with larger beams come. 

%Intro to section
This section analyzes quality degradation with the growing beam size of two systems: Neural Machine Translation (NMT) and Automatic Speech Recognition (ASR). ASR is also known to suffer from the beam-search problem~\citep{renormalize-prob-pruning-beam-asr, towards-better-decoding-asr}.
The models, training, and evaluation processes of NMT and ASR models in our work are almost identical. However, the ASR dataset (Librispeech) has some properties that are differ from the Machine Translation setting: the average length of target sentences in the training dataset is much larger than in the test, which is why we chose this task for comparison. 

%This difference helped us understand that data is the source of models biased towards short sentences. 

\subsection{Experimental setup}
\label{sec:experimental_setup_1}

%We tried to make the learning conditions of NMT and ASR as equal as possible in this section.
In order to make an informative comparative analysis of beam-search quality degradation between NMT and ASR, we aimed to make the model and experimental setup for both tasks as similar as possible to minimize mismatch. Specifically, the vocabulary, pre-processing, and models (except the first two layers of the ASR encoder) are identical between the two tasks. 

% Describe datasets, lowercase, moses and bpe
\textbf{Datasets and preprocessing}

We use IWSLT$2017$ Fr-En, IWSLT$2015$ En-Vi, WMT$2014$ En-De, and Librispeech~\citep{librispeech} datasets. The information about them is summarized in Table~\ref{tab:datasets}. We use standard validation and test splits for WMT, for the IWSLT En-Vi pair we used test $2012$ for validation and test $2013$ as a test set, for the IWSLT En-Fr pair we used development set $2010$ for validation and test $2015$ as a test set. The bulk of the analysis is done on IWSLT$17$ Fr-En and LibriSpeech-clean, as they are similar in the number of target-side tokens.  As there is no information about the case in Librispeech, we converted Librispeech and IWSLT$2017$ Fr-En to lowercase to have similar conditions for these datasets. For all other datasets the casing is unchanged. We preprocess all datasets with the Moses toolkit\footnote{\url{https://github.com/moses-smt/mosesdecoder}}, and use BPE~\citep{sennrich-etal-2016-bpe} with vocabulary size $32$k for WMT and $5$k for other datasets, as small vocabularies are beneficial for small datasets~\citep{bpe-size-2019}.

%We tried to make learning conditions between IWSLT$2017$ Fr-En and Librispeech-clean as equal as possible in order to compare them. This two datasets have approximately  As Librispeech has no information about case, we converted IWSLT$2017$ Fr-En and Librispeech to lowercase. 

% describe transformer base and ASR architecture
\textbf{Model and Optimization}

For NMT, we use Transformer base~\citep{Attentionallyouneed} model from fairseq~\citep{fairseq}. For IWSLT, we use the batch size of $8$k tokens and dropout coefficient $0.2$; all other parameters are kept as in~\citep{Attentionallyouneed}. Models are trained until convergence on a validation dataset.

For ASR, we used Transformer-base with two additional convolutional layers in the encoder, as suggested in ~\citep{fairseq-s2t}, all parameters for ASR are kept as in the original paper. 

%\begin{table}[H]
%    \centering
%    \begin{tabular}{cc}
%    \bottomrule
%    dataset & IWSLT\\
%    \midrule
%    batch size & $4$k~x~$2$GPU \\
%    dropout & $0.2$ \\
%    warmup (steps) & $16$k\\ 
%    \bottomrule
%    \end{tabular}
%    \caption{Optimization parameters for the Transformer-base model.}
%    \label{tab:optimization}
%\end{table}

\textbf{Inference and Evaluation}
\label{sec:inf_and_eval}

To produce length-normalized hypotheses, we use standard beam-search from fairseq~\citep{fairseq}. For evaluation, we averaged the last $5$ checkpoints and use BLEU~\citep{BLEU-papineni-etal-2002} computed via Sacrebleu~\citep{sacrebleu}.

For evaluating the ASR system, we used word-error-rate (WER)~\citep{WER-computation} -- a standard metric that shows edit distance from generated sequence to reference.

%For IWSLT we used BPE~\citep{sennrich-etal-2016-bpe} dictionary of 5 thousand tokens, and batch size of $8k$ tokens during training. We train models till convergence on the validation set and averaged 5 last checkpoints for all models. The rest of parameters and training procedure are kept as in original papers~\citep{Attentionallyouneed, fairseq-s2t}.

\subsection{Degradation analysis}

%\begin{table}[h!]
%    \centering
%    \small
%    \begin{tabular}{lccc}
%    \toprule
%& & BLEU  & WER \\
%\midrule
%\multicolumn{2}{l}{\bf IWSLT17}\\
%& Fr-En & 38.74 & -\\
%\multicolumn{3}{l}{\bf Librispeech}\\
%& clean & - & 6.25 (clean)\\
%& all & - & 4.24 (clean)\\
%    \bottomrule
%    \end{tabular}
%    \caption{BLEU and WER scores of our Librispeech and IWSLT2017 Fr-En models. }
%    \label{tab:nmt_vs_asr_scores}
%\end{table}

\begin{figure}[h!]
\centering
\subfigure[IWSLT$17$ Fr-En]{\includegraphics[width=0.35\textwidth]{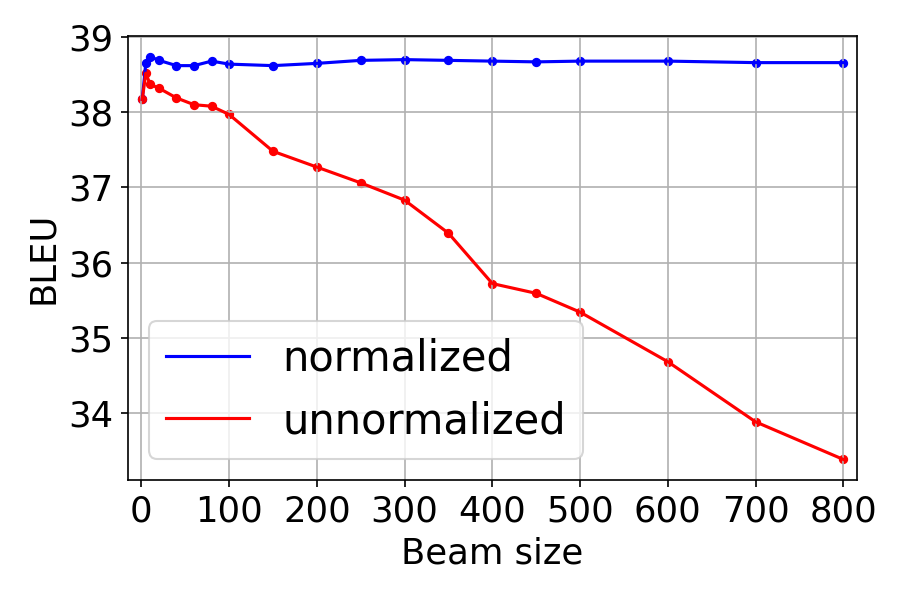}}
\subfigure[Librispeech]{\includegraphics[width=0.35\textwidth]{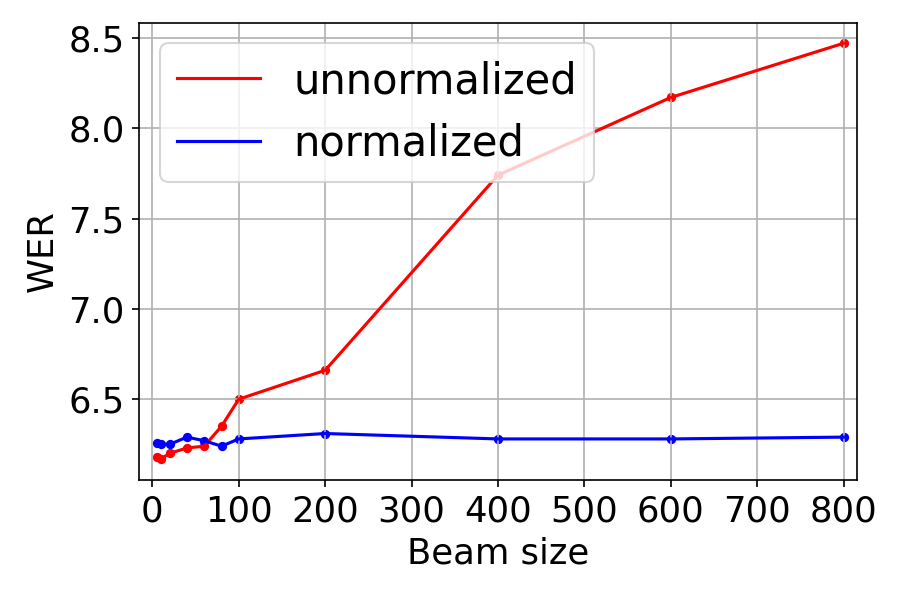}}
\caption{Comparison of BLEU and WER scores of hypotheses normalized and not normalized by length for different beam sizes. The Librispeech model was trained and tested on the 'clean' subset. }\label{fig:norm_vs_unnorm_beams}
\end{figure}

% Quality with growing beam size
In this section, we analyse which errors contribute to quality degradation with growing beam size in ASR and NMT and provide additional evidence to connect the beam-search quality degradation with the hypotheses shortening on large beams. 

Here, and later in this work, we abbreviate models with length-normalized beam search as \textit{normalized}, while models without length-normalization as \textit{unnormalized}. Figure~\ref{fig:norm_vs_unnorm_beams} shows quality of IWSLT Fr-En and Librispeech models with beam size growing from $1$ to $800$ on test sets. Normalized models do not show quality degradation with growing beam size. However, without length normalization, quality drops significantly with increasing beam-size.

%Define categories
To show which test samples cause a drop in quality, we divide the test sets into several categories, based on hypotheses from beam size $5$ and beam size $400$. These categories are following: 
\begin{itemize}
    \item b400 $\geq$ b5 -- sentences on which sentence-level BLEU of a top hypothesis from beam size $400$ is greater or equal than sentence-level BLEU of a correspondent hypothesis from beam size $5$. In other words, all cases where quality improved or didn't change with the large beam size. 
    %\item b400 $\prec$ b5 -- hypothesis from beam 400 is a full prefix of a corresponding hypothesis from beam 5 (except EOS token).
    \item b400 $\precsim$ b5 -- sentences where best hypothesis from beam $400$ is a prefix of a corresponding best hypothesis from beam $5$ (except EOS token and "." before EOS). An example of this category is a pair of hypotheses: "I can" from beam size $400$, and "I can do this tomorrow if you wait." from beam size $5$, the first is a prefix of the second;
    %except last token (which is always "." in IWSLT), b5 hypothesis from beam 400 is a full prefix of a corresponding hypothesis from beam 5 (except EOS token).
    \item b400 $<$ b5 -- all other cases that are not in the first $2$ categories. In other words, examples where quality drops, however top hypothesis from beam $400$ is not a prefix of a correspondent top hypothesis from beam $5$.
\end{itemize}

\begin{table}[h!]
\centering
\small
\begin{tabular}{l|cccc}
\toprule
Dataset & \multicolumn{2}{c}{\bf IWSLT17} & \multicolumn{2}{c}{\bf Librispeech} \\
\midrule
Subset  & unnorm. & norm. & unnorm. & norm. \\
\midrule
b400 $\geq$ b5 & 90 & 95${.}$6 & 97${.}$6 & 98${.}$7 \\
%b400 $\prec$ b5 & 0${.}$9 & 0 & 0${.}$9 & 0 \\
b400 $\precsim$ b5 & 3 & 0 & 0{.}9 & 0 \\ 
b400 $<$ b5 & 7 & 4${.}$4 & 1${.}$5 &1${.}$3 \\
\bottomrule
\end{tabular}
\caption{Distribution of cases (\%) according to defined categories between beam $5$ and beam $400$. Norm. means length-normalized versions of translations.}
\label{tab:different_drops_percents}
\end{table}

%"same or better" -- is the set of sentences on which sentence-level BLEU for beam size 400 is greater or equal to correspondent scores with beam size 5, "prefix + EOS" is where best hypothesis from beam 400 is a prefix of the corresponding hypothesis from beam 5, "prefix + . + EOS" is where best hypothesis from larger beam is the same as prefix from beam 5 hypotheses except last 2 tokens "." and "EOS" (interestingly, all hypotheses that are prefixes except two last symbols have "." before "EOS"), "other" are sentences on what we observe drop in quality and they are not included in previous groups (reformulations, synonyms for some words, grammatic errors, etc.).

% How quality degradation distributed among categories
Table~\ref{tab:different_drops_percents} shows how hypotheses are distributed among categories. The smallest is the category with prefixes -- "b400 $\precsim$ b5". Such examples are only $3\%$ of cases in IWSLT and nearly $1\%$ in Librispeech in unnormalized versions. This is significantly less than the category "b400 $<$ b5" which represents all other cases where quality drops.

\begin{table}[h!]
\centering
\small
\begin{tabular}{llccc}
\toprule
&Subset& beam 5 & beam 400 & Contribution to \\
&&&&degradation\\
\midrule
\multicolumn{3}{l}{\bf IWSLT Fr $\rightarrow$ En, BLEU} &&\\
&b400 $\geq$ b5     & 38.83     & 39.29 & {+}0.41 \\
&b400 $\precsim$ b5 & {\bf 38.71 }    & {\bf 0.13} & {\bf -1.16 }  \\ 
&b400 $<$ b5        & 36.70    & 30.75  & -0.42    \\ 
%&whole test      & 38.54        & 35.74        & -2.8    \\ 
\midrule
\multicolumn{3}{l}{\bf Librispeech, WER} &&\\
&b400 $\geq$ b5 & 5{.}5 & 5{.}4 & -0.1 \\
&b400 $\precsim$ b5 & {\bf22{.}9  }  &  \textbf{86{.}5} & \textbf{+0.57}  \\ 
&b400 $<$ b5     & 17{.}6      & 26{.}6  & +0.13    \\ 
%whole test      & 6{.}2        & 7{.}7        & +1{.}5\\ 
\bottomrule
\end{tabular}
\caption{BLEU scores for IWSLT$17$ and WER scores for Librispeech corresponding to the different categories. Column "Contribution to degradation" represents the difference between beam $400$ and beam $5$ scores, weighted by the percentage of the dataset in the corresponding category from Table~\ref{tab:different_drops_percents}.}
\label{tab:quality_for_categories}
\end{table}

Although examples where the EOS token appeared too early during the generation of a reasonable, long hypothesis, are \emph{smallest} category, they have the \emph{greatest} contribution to the overall quality degradation with growing beam size. Consider Table~\ref{tab:quality_for_categories}, which shows quality in terms of BLEU/WER for different categories and beam sizes. The biggest drop in quality is in the prefixes category. It drops from $38.71$ BLEU to almost $0$ for IWSLT. For Librispeech, WER increases from $22.9$ to $86.5$ in the same category. Performance within the category "b400 $<$ b5" degrades more modestly, losing $5$ BLEU and gaining $3.52$ WER, respectively. Weighed by the fraction of each category within the datasets, the prefixes category contributes nearly $3$ times more than the non-prefix category to the \emph{overall} BLEU on IWSLT ($1.16$ vs. $0.42$), and nearly $4$ times more to the \emph{overall} WER on Librispeech ($0.57$ vs. $0.13$)\footnote{This is an approximate analysis, as BLEU is a non-linear corpus-level statistic.}.
\begin{table}[h!]
\centering
\small
\begin{tabular}{lcccc}
\toprule
&Subset & beam 5 & beam 400 & Contribution to \\
&&&& shortening \\
\midrule
\multicolumn{3}{l}{\bf IWSLT Fr $\rightarrow$ En} \\
&b400 $\geq$ b5 & 22.02 & 21.69 & -0.3 \\
&b400 $\precsim$ b5 & \textbf{53.8} &	{\bf 8.72}&	{\bf-1.35} \\
&b400 $<$ b5     &44.2&	39.3& -0.34    \\
&whole test  &24.53&	22.54&	-1.99   \\
\midrule
\multicolumn{3}{l}{\bf Librispeech} \\
&b400 $\geq$ b5 & 26.12 & 26.11 & -0.01 \\
&b400 $\precsim$ b5 & {\bf 67.17 }&	{\bf 10.65}& {\bf -0.51}	 \\
&b400 $<$ b5 &42.67&	39.15&  -0.05  \\
&whole test  & 26.73 &	26.17 &	-0.56   \\
\bottomrule
\end{tabular}
\caption{Average token lengths of best hypotheses from different beam sizes and categories. Column "Contribution to shortening" represents the difference between columns "beam $400$" and "beam $5$" weighted by the fraction of the corresponding category in the dataset.}
\label{tab:lengths_for_categories}
\end{table}

%Say that lengths mostly decrease also from prefixes
Table~\ref{tab:lengths_for_categories} shows that the prefix category (early EOS) is also the most significant in terms of length reduction with growing beam size. Length for beam $400$ in this category is nearly $84\%$ lower compared to beam $5$. Interestingly, early EOS occurs mainly in examples where the top hypothesis from beam $5$ is long, on average $53{.}8$ tokens in IWSLT and $67{.}17$ in Librispeech, which is much longer than average lengths over the whole test datasets, $24{.}53$ and $26{.}73$ respectively. This observation adds further evidence to the connection between hypotheses shortening and the quality degradation with growing beam size. 

Our findings relate to work studying calibration\footnote{Calibration measures how well the model's probability distribution matches the actual data distribution.} problems of NMT, which show that NMT architectures are poorly calibrated, especially the EOS token~\citep{calibration-of-encoder-decoder,inference-calib-2020}.

\subsection{Dataset length-bias}

\begin{figure*}[!ht]
    \centering
    \subfigure[IWSLT Fr-En]{\includegraphics[width=0.24\textwidth]{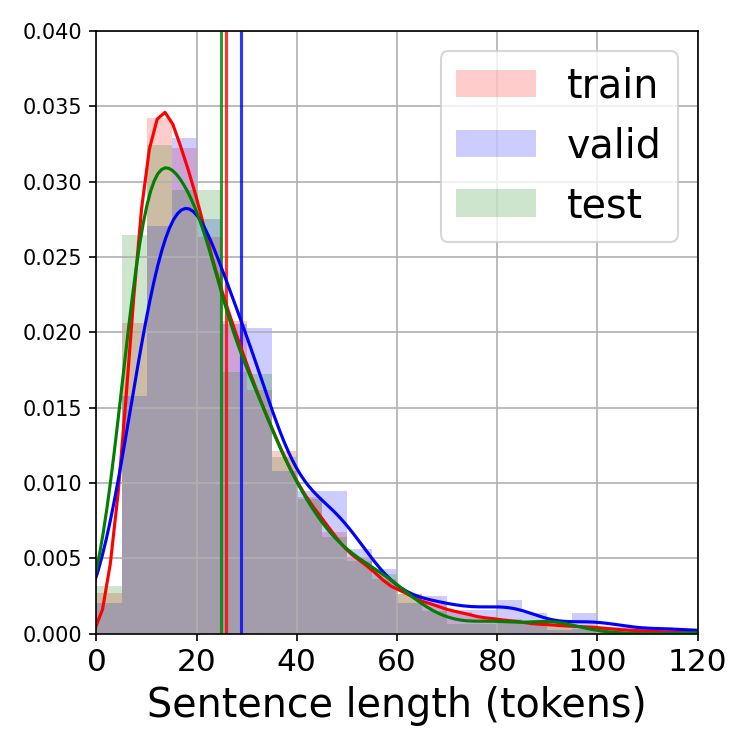}}
    \subfigure[IWSLT En-Vi]{\includegraphics[width=0.24\textwidth]{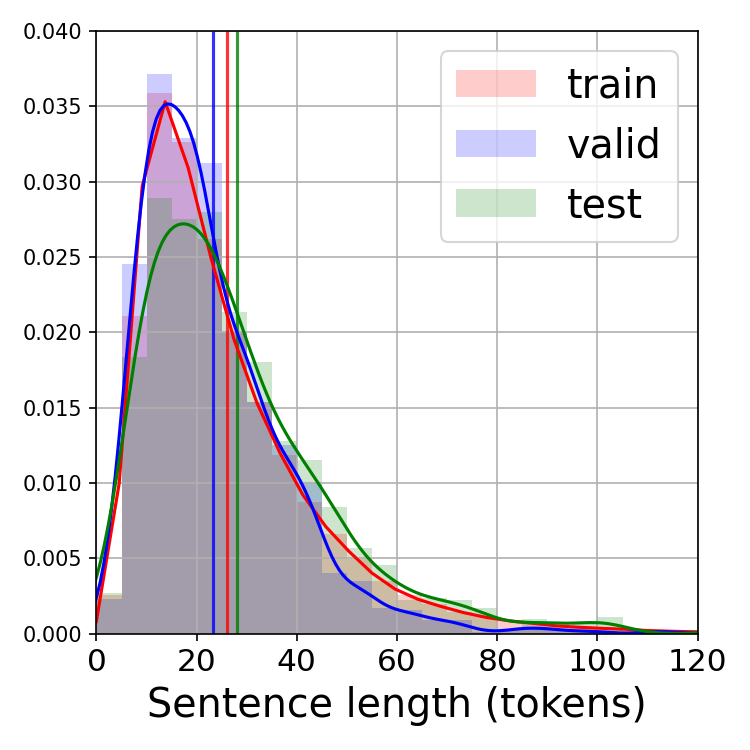}}
    \subfigure[WMT De-En]{\includegraphics[width=0.24\textwidth]{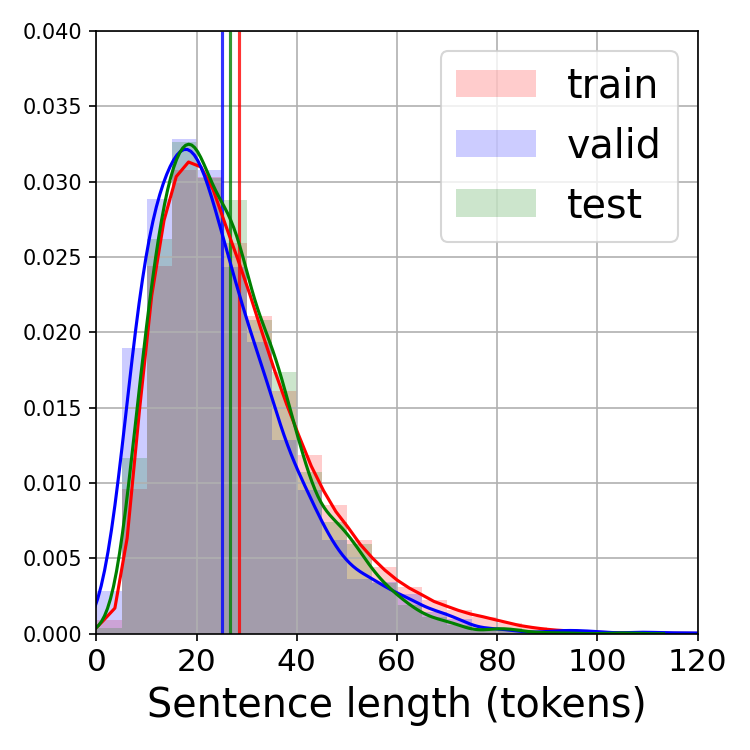}}
    \subfigure[Librispeech]{\includegraphics[width=0.24\textwidth]{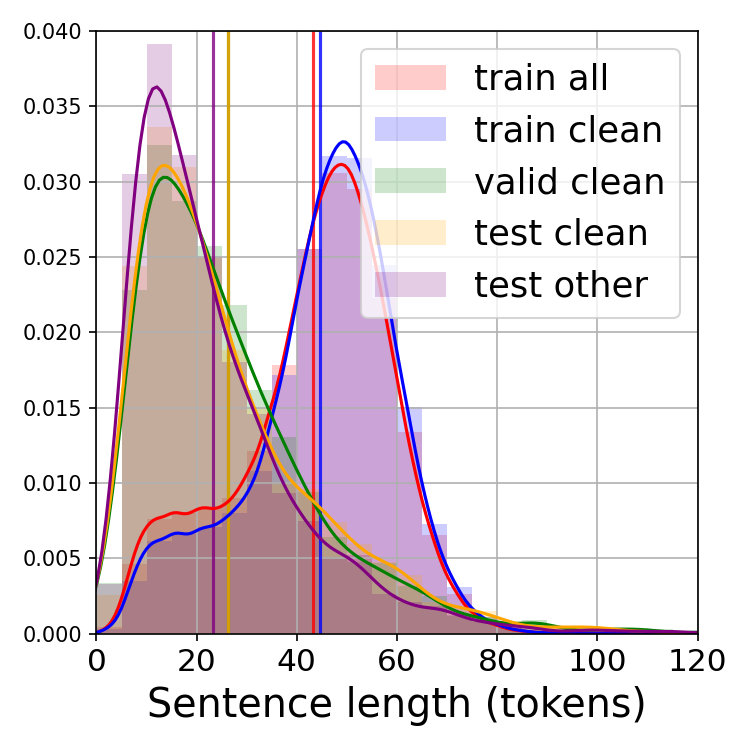}}
    \caption{Distribution of target side sentence lengths after Moses and BPE for different datasets. Vertical lines represent mean of the corresponding distribution.}
    \label{fig:dataset_token_lengths_dist}
    
    \subfigure[IWSLT Fr-En BLEU]{\includegraphics[width=0.24\textwidth]{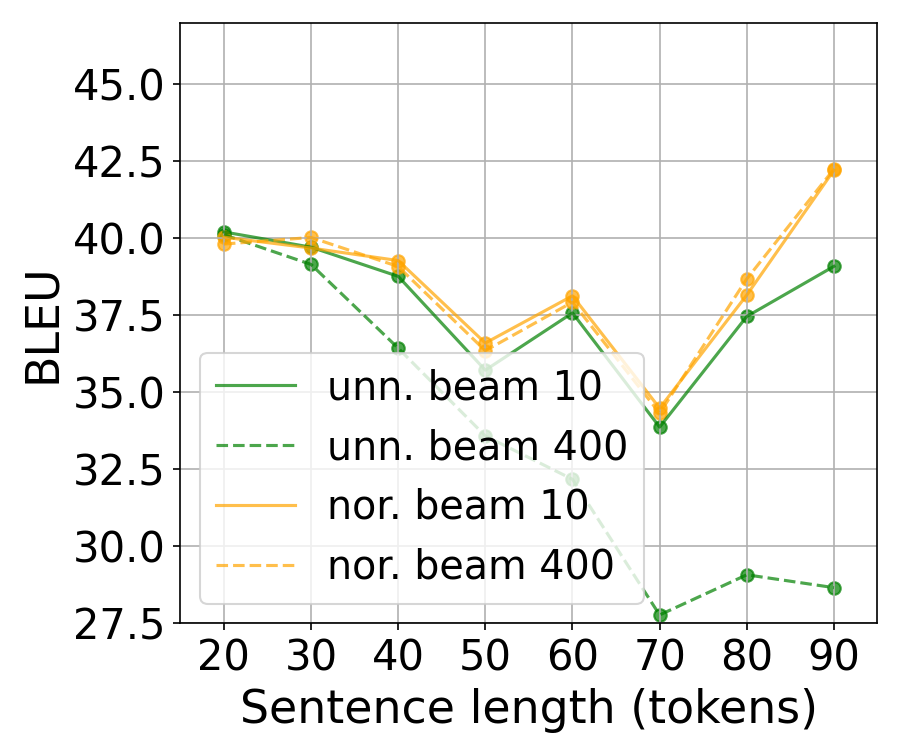}}
    \subfigure[IWSLT En-Vi BLEU]{\includegraphics[width=0.24\textwidth]{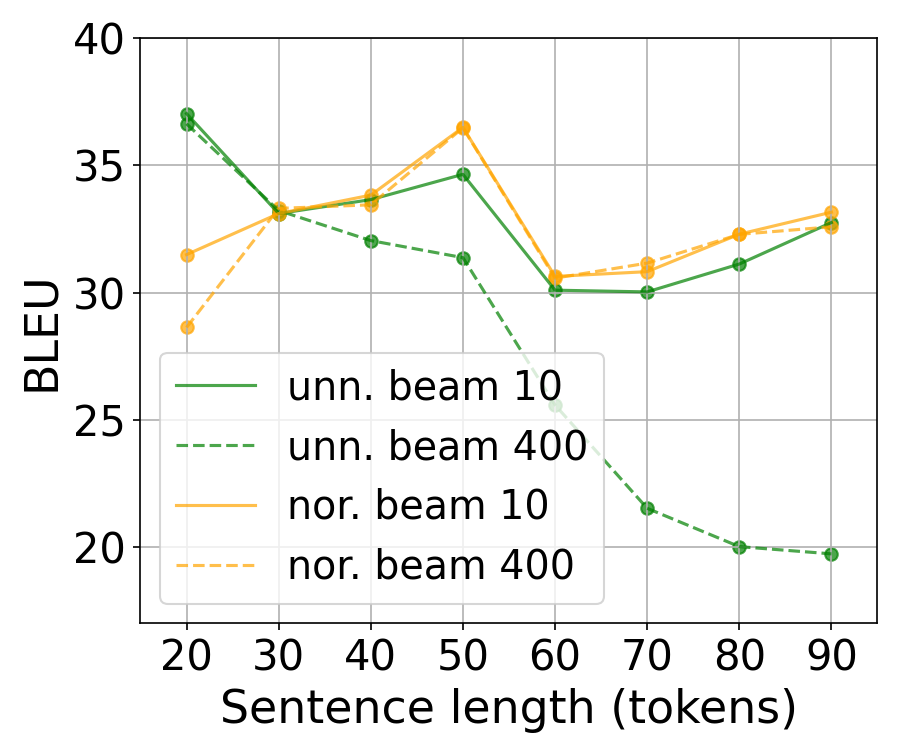}}
    \subfigure[WMT De-En BLEU]{\includegraphics[width=0.24\textwidth]{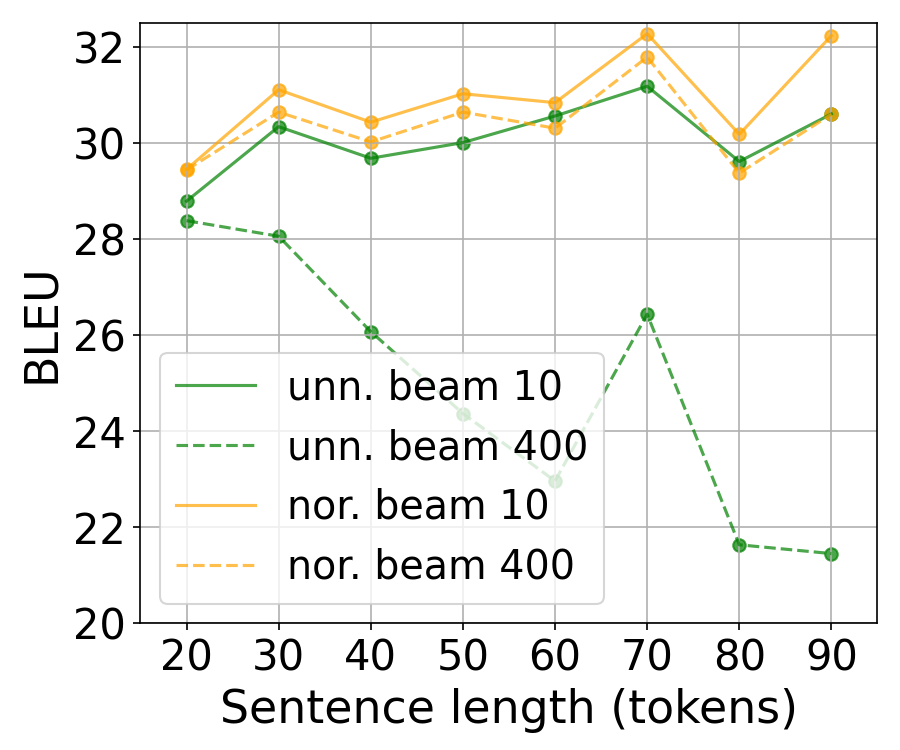}}
    \subfigure[Librispeech-clean WER]{\includegraphics[width=0.24\textwidth]{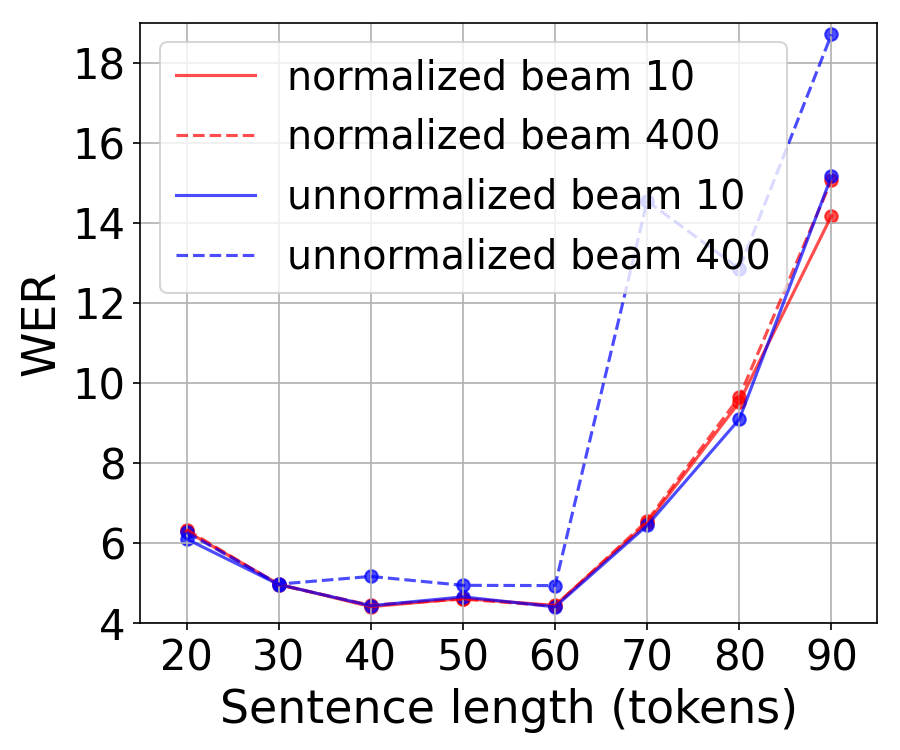}}
%    \subfigure[IWSLT MSR N=2]{\includegraphics[width=0.3\textwidth]{imgs/iwslt_en_MSR_2_bleu_by_length_buckets.png}}
%    \subfigure[IWSLT MSR N=3]{\includegraphics[width=0.3\textwidth]{imgs/iwslt_en_MSR_3_bleu_by_length_buckets.png}}
%    \subfigure[IWSLT MSR N=4]{\includegraphics[width=0.3\textwidth]{imgs/iwslt_en_MSR_4_bleu_by_length_buckets.png}}
    \caption{BLEU scores for buckets with growing sentence length. Each point represents quality calculated on sentences with lengths from the previous point to the current position, starting from $0$.}
    \label{fig:default_models_buckets}
\end{figure*}

Having found further evidence to link length bias with the beam search problem, we examine and compare the distribution of sentence lengths in typical NMT and ASR datasets.

% Show correlation between quality degradation and growing beam size, emphasize Librispeech behaviour
Consider Figure~\ref{fig:dataset_token_lengths_dist}, which shows the distribution of sentence lengths in IWSLT$17$ Fr-En, IWSLT$15$ En-Vi, WMT$14$ De-En, and Librispeech datasets. The NMT datasets have an average sentence length between $20$ and $25$ tokens and exhibit a strong, asymmetric skew towards short sentences. In contrast, the Librispeech \emph{training} dataset exhibits a more symmetric distribution of sentence length, with an average length of more than $40$. At the same time, the distribution of lengths in the test and validation sets is similar to NMT. As a result, \textit{during training on Librispeech, the model sees a far larger and diverse set of long sentences than is encountered in evaluation}. 

Let's investigate how quality relates to the length of the target sentences. Figure~\ref{fig:default_models_buckets} shows BLEU/WER scores on the test sets for buckets based on the reference length. If we compare Figure~\ref{fig:dataset_token_lengths_dist} with Figure~\ref{fig:default_models_buckets} we will note an interesting feature: quality degradation on unnormalized beam $400$ in machine translation tasks starts after length around $30$ and mostly monotonically increases as we go to longer sentences. In contrast, on Librispeech, the quality starts to drop only after the length $60$, which correlates with the distribution of lengths of train examples. Specifically, in IWSLT$17$ Fr-En dataset, only $30\%$ of training sentences have lengths greater than $30$ and less than $10\%$ are longer than $50$ tokens. In contrast, Librispeech has many training sentences with a length of $60$ tokens or less, and their amount drops rapidly only after this value, with about $5\%$ of sentences having a length greater than $65$ tokens.

Thus, we can clearly see that beam search quality degrades when operating on sentences, which are underrepresented in the training datasets in terms of \emph{reference} length. This brings us to one of the main ideas of our work: \textit{training datasets biased towards short sentences strongly contribute to the quality degradation with growing beam size}. In typical training datasets in Machine Translation (IWSLT and WMT) long sentences are significantly underrepresented, causing models to overfit to shorter sentences and overestimate probabilities of short prefixes. In the next section, we propose a data-augmentation strategy to alleviate this issue.

\section{Multi-Sentence Resampling}

%\todo{In this section we introduce our new method, show that it solves problem of beam degradation (and prefixes?), show score gaines obtained with it on different datasets}

%In this section we introduce \textit{Multi-Sentence Resampling} -- a simple data augmentation method that solves problem of length bias and the degradation of quality with growing beam size. 

%As we show in the previous section, simple oversampling, when we make many copies of the same small set of long sentences does not work well: the resulting quality does not improve and continues to drop with growing beam size. Likely because the model overfits for specific examples.
%To solve this issue, we propose to concatenate random train sentences one after another in order to significantly increase length of train examples without copying same long sentences many times. As our results show, this method works surprisingly well. 

\begin{figure*}[h!]
    \centering
    \includegraphics[width=0.95\textwidth]{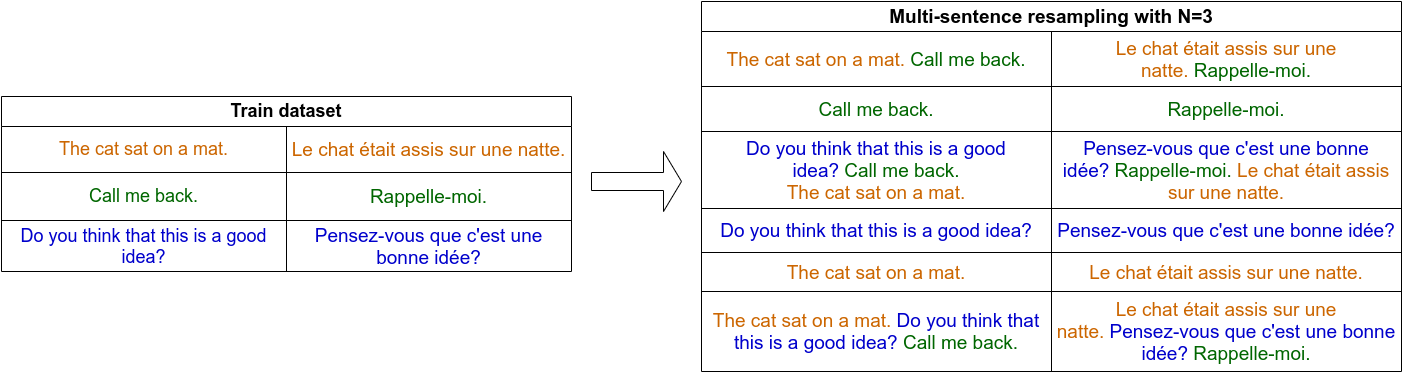}
    \caption{Example of Multi-Sentence Resampling with N=3. In the new train dataset with equal probability there are examples with $1$,$2$ and $3$ sentences from the original dataset.}
    \label{fig:msr_describtion}
\end{figure*}

\begin{algorithm}
 \vspace{2px}
 \hrule
 \vspace{3px}
 {\textbf{Algorithm 1: Multi-Sentence Resampling}} \vspace{3px} \hrule \vspace{3px}
    $D \leftarrow$ old train dataset with pairs of sentences;\\
    $N \leftarrow$ maximum number of sentences in one example in a new train dataset;\\
    $S = |D| \cdot M \leftarrow $ number of examples in the new train dataset;\\
    $R \leftarrow \emptyset$ -- new dataset;\\
    \For{$i$ in $1..S$} {
        $n \leftarrow $ random integer from $1$ to $N$ \\
        new\_source $=$ ""\\
        new\_target $=$ ""\\
        \For{$k$ in $1..n$} {
            (cur\_source, cur\_target) $=$ sample\ random\ example\ from\ $D$\\
            new\_source $\mathrel{{+}{=}}$ cur\_source \\
            new\_target $\mathrel{{+}{=}}$ cur\_target \\
        }
        $R.append(($new\_source, new\_target$))$\\
        } 
    \Return{$R$}\;
 \hrule
\end{algorithm}
In this section, we introduce \textit{Multi-Sentence Resampling} (MSR) -- a simple data augmentation method that alleviates the beam search problem by addressing dataset length bias and which increases the overall quality of translation models. Specifically, MSR augments a dataset such that instead of one sentence, each training example consists of $1$ to $N$ sentences, randomly chosen from a dataset and concatenated one after another. It preserves the order of sentences: the source side is concatenated to the source side, and the target side is concatenated to the target side of a new train example. For each new training example, MSR randomly chooses from $1$ to $N$ sentences, so that the model does not overfit to the particular number of sentences.

Figure~\ref{fig:msr_describtion} illustrates the algorithm. The original training dataset, with $3$ examples, is on the left, and the dataset created by \textit{Multi-Sentence Resampling} is shown on the right. The new dataset contains $6$ train examples: $2$ examples with one sentence, $2$ examples with a concatenation of two original sentences, and $2$ examples with a concatenation of $3$ original sentences. The full procedure is described in Algorithm~1. The algorithm takes an original train dataset, desired number of examples in the augmented dataset $S$, and the maximum number of sentences in an example $N$. \textit{Multi-Sentence Resampling} increases the average length of the dataset and provides a diverse set of long training examples. In contrast to other methods that use rescoring of hypotheses and per-token rewards~\citep{breaking-beam-curse-2019} or predict target length separately~\citep{predicting-length-2020}, our method does not change the search procedure.

\begin{figure}[h]
    \centering
    \includegraphics[width=0.4\textwidth]{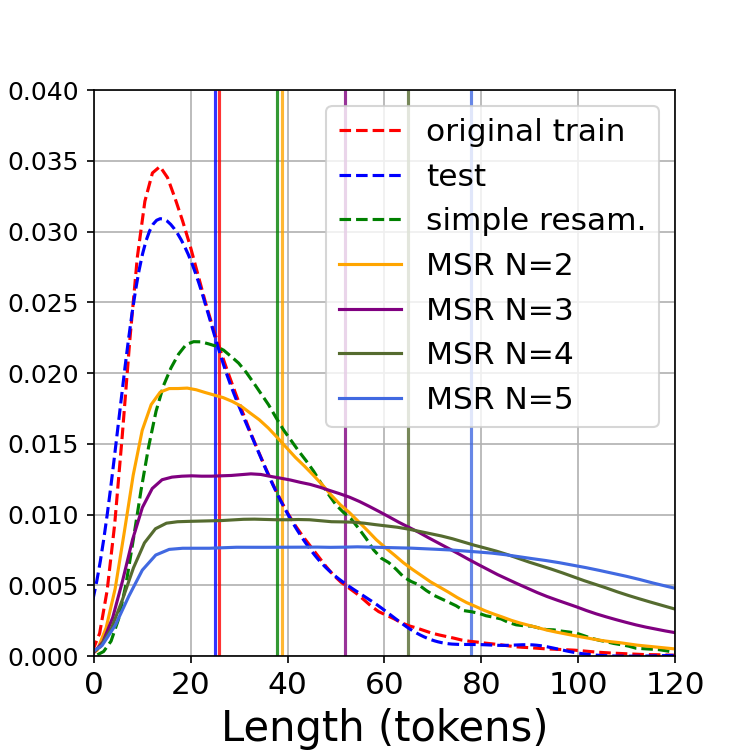}
%    \subfigure[IWSLT]{\includegraphics[width=0.3\textwidth]{imgs/iwslt_en_token_lengths.png}}
%    \subfigure[IWSLT MSR]{\includegraphics[width=0.3\textwidth]{imgs/iwslt_resampled_fr_en_token_lengths.png}}
%   \subfigure[IWSLT MSRw]{\includegraphics[width=0.3\textwidth]{imgs/iwslt_resampled_fr_en_length_normalized_token_lengths.png}}
    \caption{Distribution of sentence lengths for different $N$ in MSR for IWSLT$17$ Fr-En dataset (en part). Vertical lines represent mean of the corresponding distribution.}
    \label{fig:msrs_token_length}
\end{figure}

Figure~\ref{fig:msrs_token_length} illustrates how train examples length distribution changes in IWSLT$17$ Fr-En dataset for $N$ from $2$ to $5$. With growing $N$ distributions become more flatten for lengths presented in the test set. The average length of examples in a new train dataset can be approximately calculated as  $$new\_length\simeq \sum_{n=1}^N \frac{L \cdot n}{N} =  L\cdot \frac{N+1}{2},$$ where $L$ is the average length of the original dataset.

%In order to try to solve train length imbalance in the dataset we firstly try to  oversample long sentences. We created two datasets, first is IWSLT resampled to make a length distribution similar to Librispeech, and the second is a dataset where sentences from IWSLT are sampled with probabilities proportional to their length. Figure~\ref{fig:simpleoversampling} (a) shows distribution of lengths of train examples for each case, while~\ref{fig:simpleoversampling} (b) shows quality degradation with growing beam size. It is clear that simple oversampling only slightly reduces beam quality degradation and does not improve overall quality. In the next section we describe a dataset augmentation method that both: solves quality degradation with growing beam size problem, and improve overall quality of a model.

%\begin{figure}[h!]
%    \centering
%    \subfigure[token lengths]{\includegraphics[width=0.2\textwidth]{imgs/iwslt_en_simple_oversampling_token_lengths.png}}
    %\subfigure[growing beam]{\includegraphics[width=0.2\textwidth]{imgs/iwslt_simple_resampling_beam.png}}
    %\subfigure[buckets]{\includegraphics[width=0.3\textwidth]{imgs/librispeech_iwslt_resampled_wer_by_length_buckets.png}}
%    \caption{Simple resampling.}
%    \label{fig:simpleoversampling}
%\end{figure}

\section{Experiments}

In this section we provide an empirical evaluation of the \textit{Multi-Sentence Resampling} on the IWSLT and WMT datasets.

%\subsection{Experimental setup}
%\label{sec:experimental_setup_2}
%For IWSLT$2017$ En-Fr and IWSLT$2015$ En-Vi settings were presented previously in Section~\ref{sec:experimental_setup_1}. Here we provide additional experimental settings. 

%\subsubsection{Datasets and preprocessing}

%In this section we additionally use WMT$2014$ En-De dataset. It is described in Table~\ref{tab:datasets}. Figure~\ref{fig:wmt_token_lengths_dist} represents distribution of sentence lengths for WMT En-De. 

%\begin{figure}[h!]
%    \centering
%    \includegraphics[width=0.4\textwidth]{imgs/wmt_en_token_lengths.png}
%    \caption{Distribution of sentence lengths after tokenization and BPE for WMT$2014$ En-De. Vertical lines represent mean of the corresponding distribution.}
%    \label{fig:wmt_token_lengths_dist}
%\end{figure}

%\subsection{Model and Optimization}

%\begin{table}[h!]
%    \centering
%    \begin{tabular}{ccc}
%    \bottomrule
%    dataset & IWSLT& WMT \\
%    \midrule
%    batch size & $4$k~x~$2$GPU & $4$k~x~$8$GPU \\
%    dropout & $0.2$ & $0.1$ \\
%    warmup (steps) & $16$k & $4$k \\ 
%    fp16 & yes & yes \\ 
%    \bottomrule
%    \end{tabular}
%    \caption{Optimization parameters.}
%    \label{tab:optimization}
%\end{table}

\subsection{Experimental setup}

To compare with the original Transformer paper~\citep{Attentionallyouneed}, where authors used another beam-search parameters and BLEU computation, we added additional part to Table~\ref{tab:main_multisentence_scores}. For this part we changed length penalty to $0.6$ in beam-search and compute BLEU as in~\citep{Attentionallyouneed}\footnote{\url{https://github.com/pytorch/fairseq/blob/master/scripts/compound_split_bleu.sh}}. As baselines we use standard models trained without data augmentation.

In our experiments, for IWSLT datasets $M=10$ -- the new train dataset is $10$ times larger than the original one, for WMT$2014$ En-De $M=5$, as this dataset is much bigger.

\subsection{Quality with growing beam size}

\begin{figure}[h!]
    \centering
    \subfigure[IWSLT$17$ Fr-En]{\includegraphics[width=0.4\textwidth]{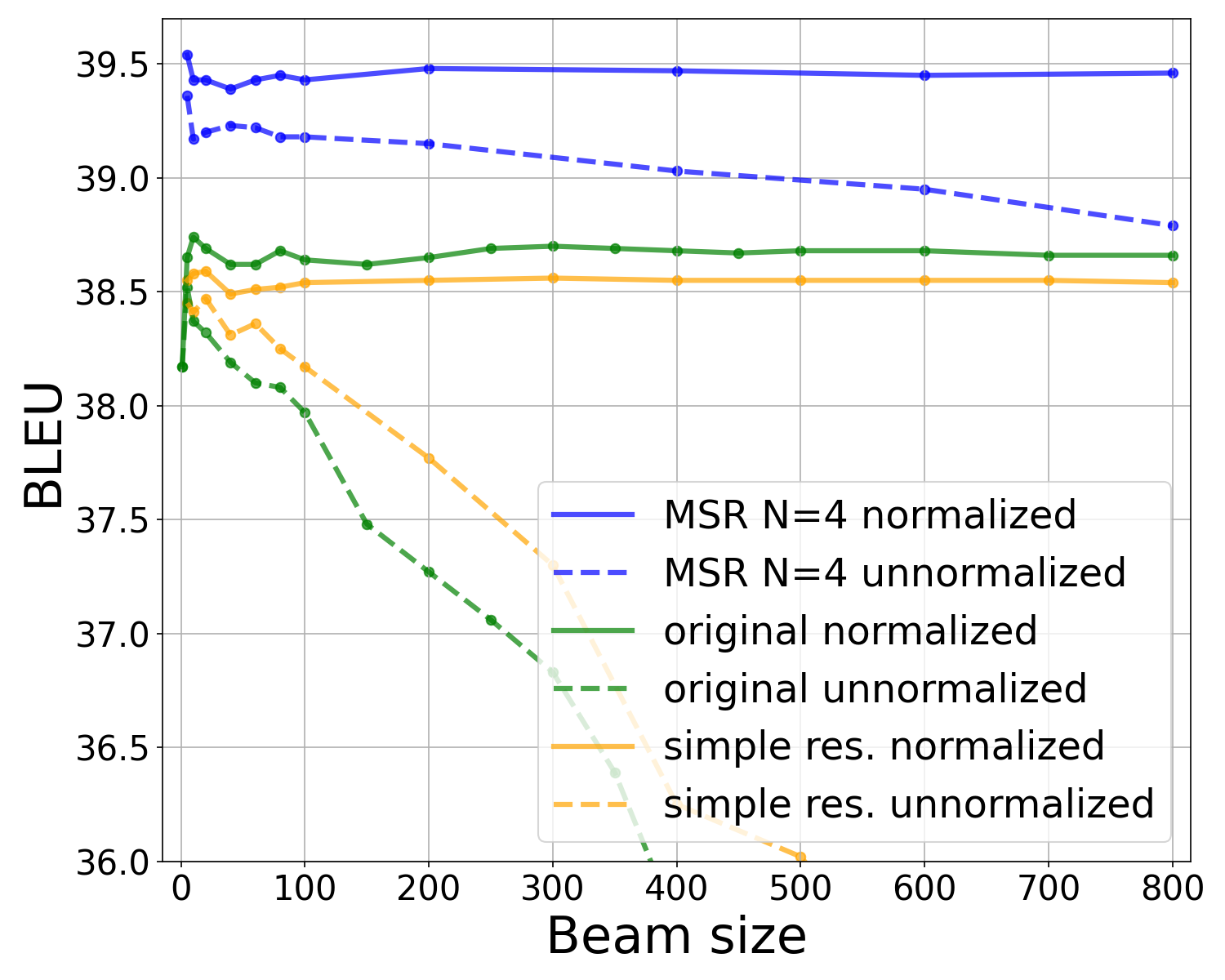}}
    \subfigure[WMT$14$ En-De]{\includegraphics[width=0.4\textwidth]{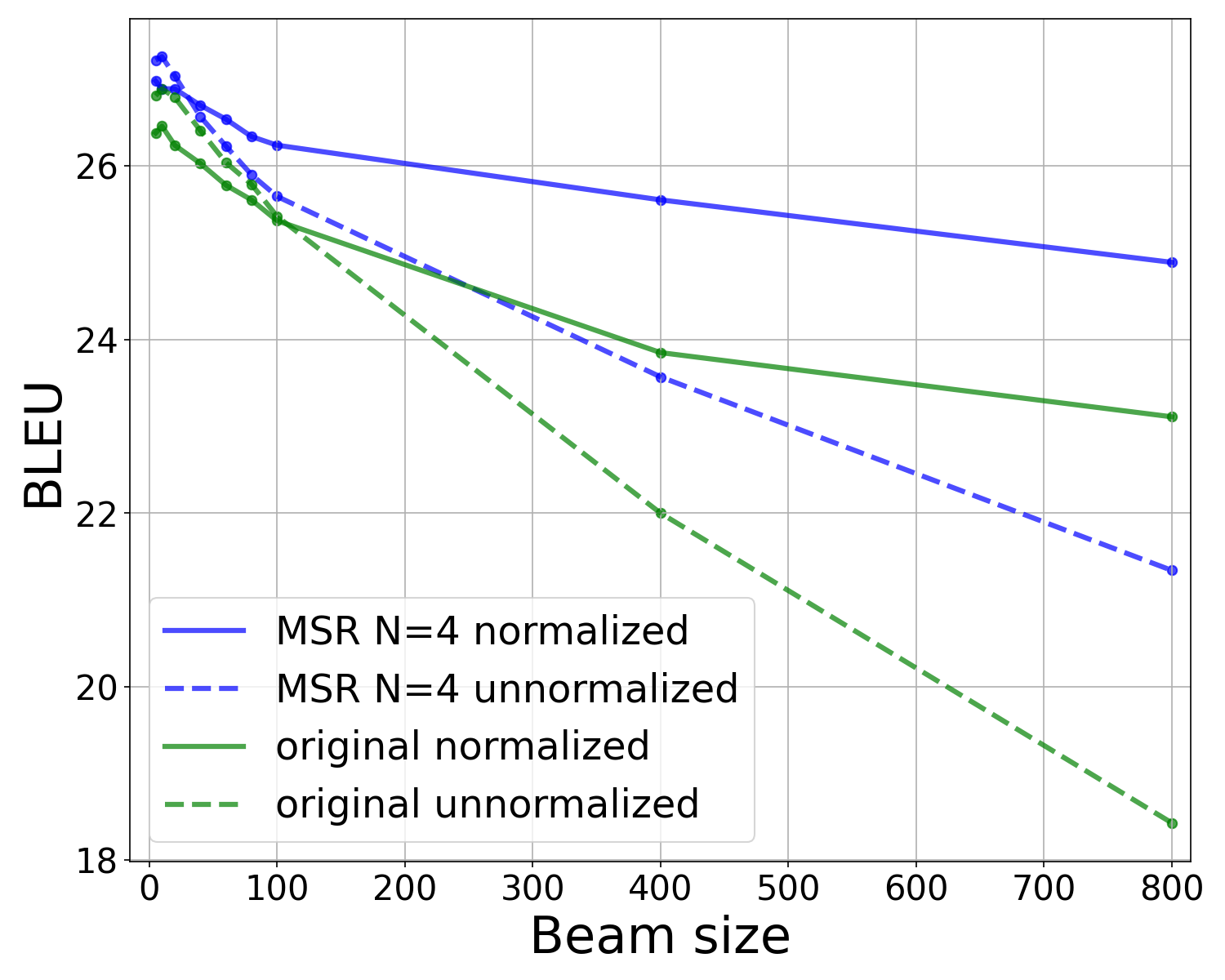}}
    %\subfigure[IWSLT normalized]{\includegraphics[width=0.4\textwidth]{imgs/MSR_beam_normalized.png}}
    %\subfigure[MSRw normalized]{\includegraphics[width=0.4\textwidth]{imgs/MSR_w_beam_normalized.png}}
    %\subfigure[IWSLT unnormalized]{\includegraphics[width=0.4\textwidth]{imgs/MSR_beam_unnormalized.png}}
    %\subfigure[WMT normalized]{\includegraphics[width=0.4\textwidth]{imgs/WMT_beam_normalized.png}}
    %\subfigure[WMT unnormalized]{\includegraphics[width=0.4\textwidth]{imgs/WMT_beam_unnormalized.png}}
    %\subfigure[MSRw unnormalized]{\includegraphics[width=0.4\textwidth]{imgs/MSR_w_beam_unnormalized.png}}
    \caption{ Quality with growing beam size of the baseline, MSR $N=4$ and simple resampling.}
    \label{fig:msr_quality_figures}
\end{figure}

Figure~\ref{fig:msr_quality_figures} compares quality degradation with growing beam size for the baseline and MSR with $N=4$ for IWSLT$17$ Fr-En datasets and WMT$14$ En-De. As an additional baseline, we compare MSR with a simpler strategy on IWSLT -- resampling the dataset multiple times so that the probability of a sentence is proportional to its length. This way, long sentences occur more frequently during training. There are several interesting points in this comparison. Firstly, Multi-Sentence Resampling achieves significantly better quality than the baseline on both datasets. Secondly, while the baseline's quality rapidly drops with the growing beam size, the quality of the \textit{Multi-Sentence Resampling} drops much more slowly. In particular, MSR with $N=4$ with beam size $800$ has quality better than the baseline with \emph{any} beam size on IWSLT. On WMT, improvements for large beam sizes are more modest, which is expected, as data augmentation has less effect on larger datasets. However, MSR works on par with the length-normalized baseline up to the beam size $400$. Third, simple resampling in IWSLT works slightly better than the baseline in the unnormalized setting; however, it drops quality in the normalized case. The benefits of simple resampling are limited because, unlike with MSR, the set of long sentences severely lacks diversity, and the model overfits to it during training. 

\begin{figure}[h!]
    \centering
    \subfigure[IWSLT unnormalized]{\includegraphics[width=0.4\textwidth]{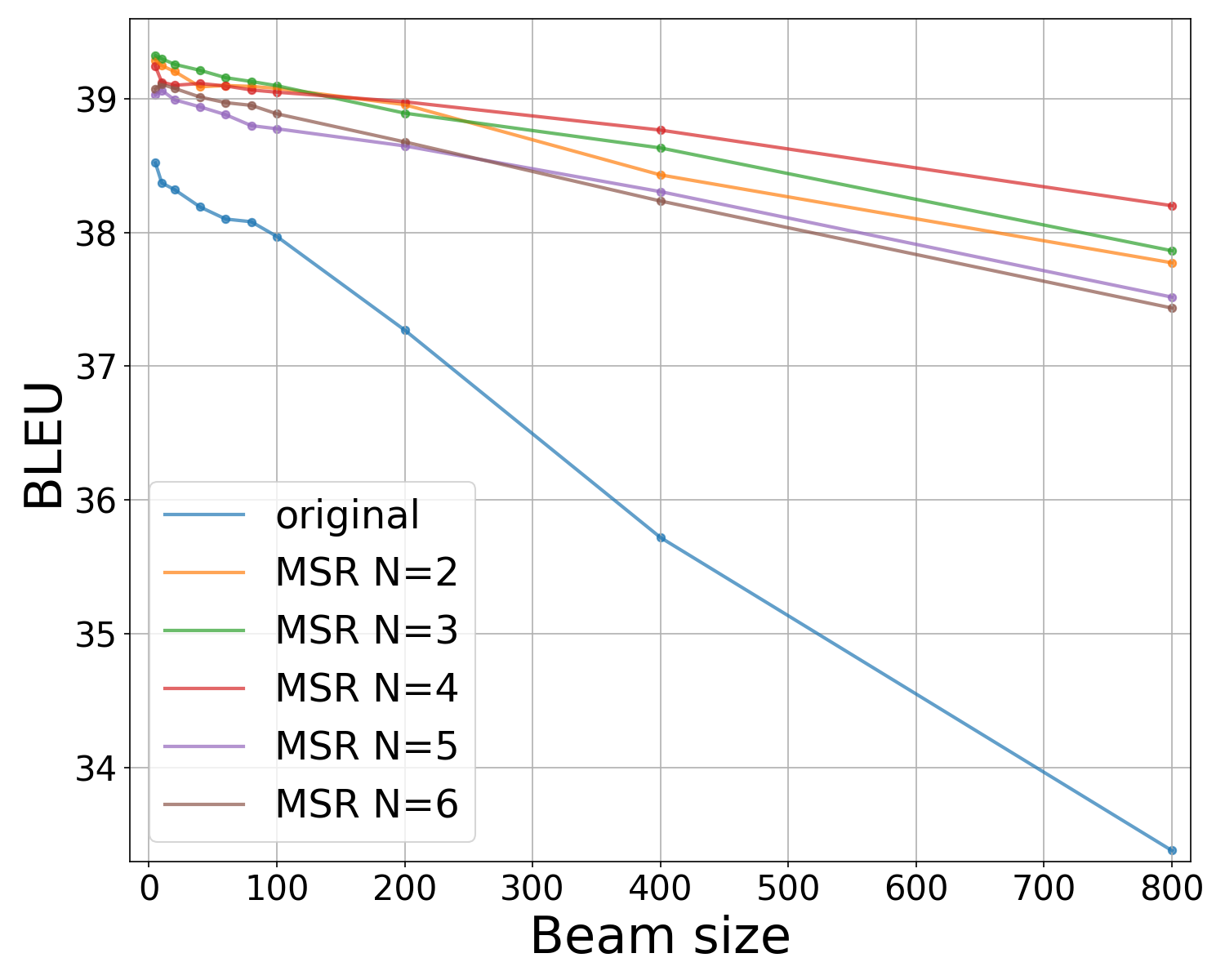}}
    % \subfigure[IWSLT unnormalized]{\includegraphics[width=0.4\textwidth]{imgs/MSR_iwslt_choice_of_N_unnormalized.png}}
    %\subfigure[MSRw unnormalized]{\includegraphics[width=0.4\textwidth]{imgs/MSR_w_beam_unnormalized.png}}
    \caption{BLEU with growing beam size for Multi-Sentence resampling with different $N$ on IWSLT2017 Fr-En dataset. Each point is an average between $3$ models trained with different random seeds.}
    \label{fig:msr_choise_of_N}
\end{figure}

We analyze how the value of the hyperparameter $N$ in MSR affects beam search quality in Figure~\ref{fig:msr_choise_of_N}, which shows the quality of trained models for different values of $N$ across a range of beam sizes.  On small beams, all $N$ behave without significant difference. On large beams, quality grows with $N$ up to $4$ and decreases for larger $N$.  

\begin{figure*}[!ht]
    \centering
    \subfigure[Original]{\includegraphics[width=0.24\textwidth]{imgs/iwslt_nmt_default_by_length_buckets.png}}
    \subfigure[MSR N=2]{\includegraphics[width=0.24\textwidth]{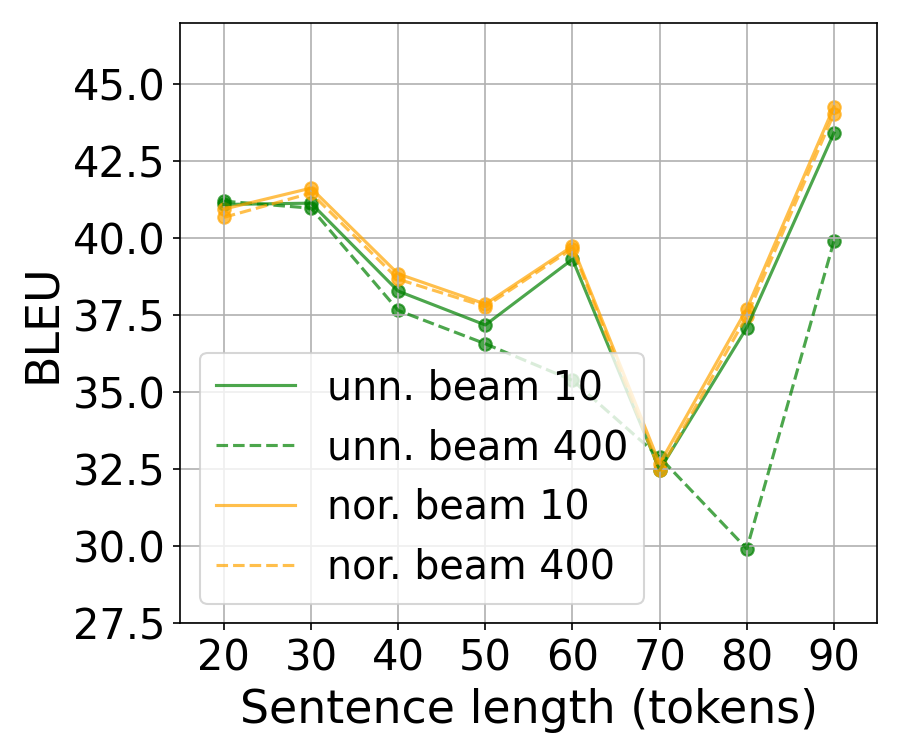}}
    \subfigure[MSR N=3]{\includegraphics[width=0.24\textwidth]{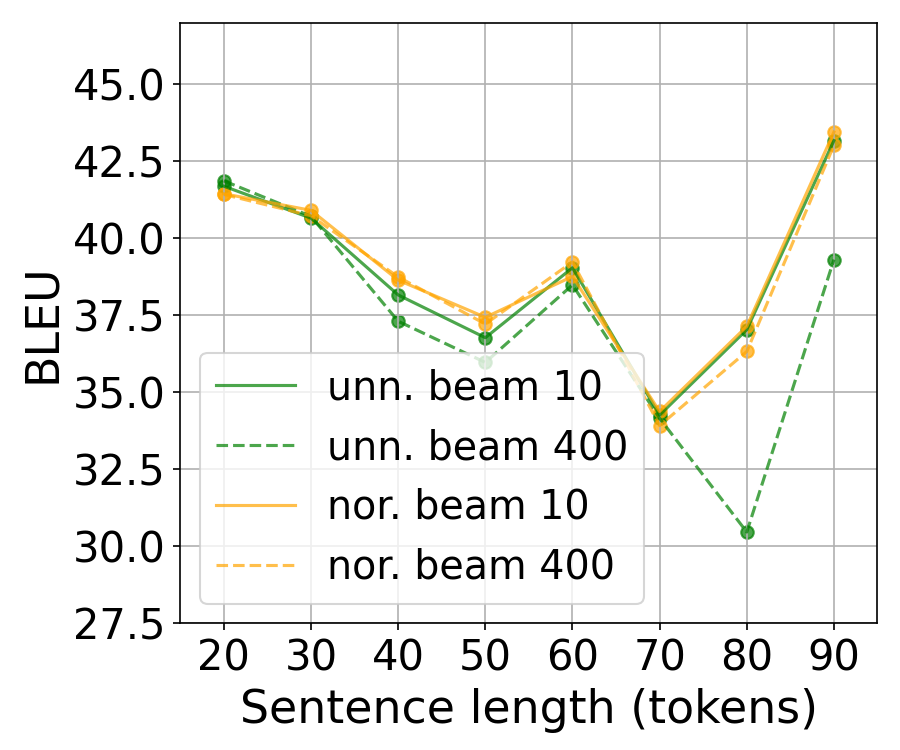}}
    \subfigure[MSR N=4]{\includegraphics[width=0.24\textwidth]{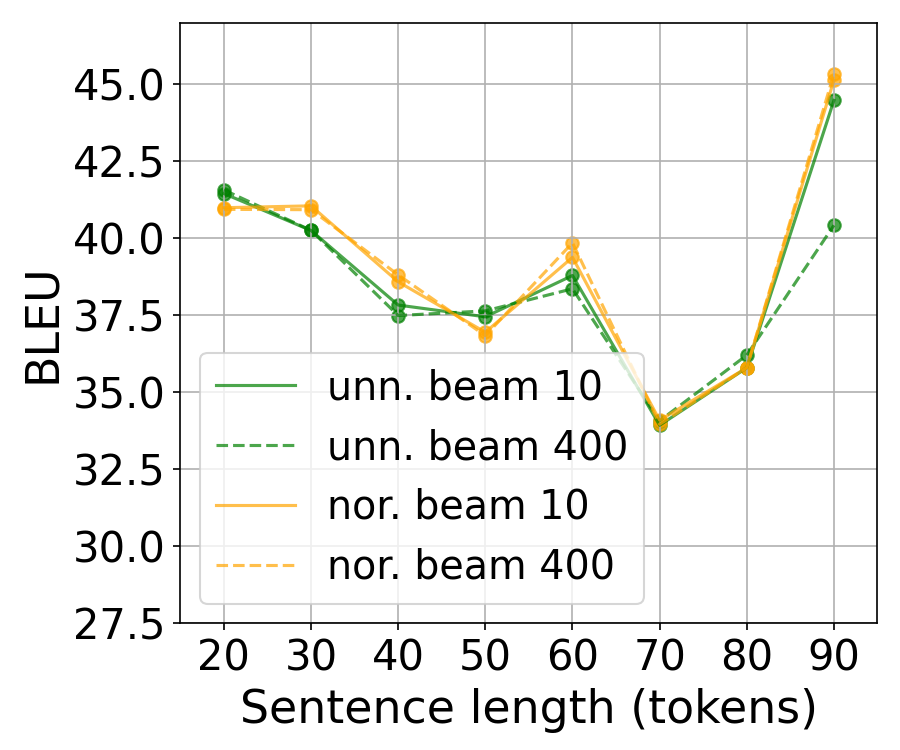}}
    \caption{Changes of quality with growing sentence length for IWSLT$17$ Fr-En. Each point represents quality calculated on sentences with lengths from previous point to the current position, starting from $0$.}
    \label{fig:msr_quality_buckets}
\end{figure*}

An additional analysis of how effects of the \textit{Multi-Sentence Resampling} vary with reference length and number of concatenated sentences is provided in Figure~\ref{fig:msr_quality_buckets}. Baseline unnormalized beam-search with beam-width $400$ works badly on long sentences: quality degradation increases starting from length $30$. On the other hand, MSR with $N=4$ has almost no degradation for long sentences. Additionally, experiments with $N=2$ and $N=3$ show that quality degradation on long sentences decreases as we increase $N$, likely because we start fitting the model to far longer sequence lengths than encountered in the test set.

\begin{table}[h!]
\centering
\begin{tabular}{lccccc}
\toprule
& & \multicolumn{2}{c}{\bf original}  & \multicolumn{2}{c}{\bf MSR} \\
& & norm. & unn. & norm. & unn. \\
\midrule
\multicolumn{3}{l}{\bf IWSLT17}\\
& Fr-En & 38{.}74 & 38{.}37 & \bf{39{.}4} & \bf{39{.}23} \\
& En-Fr & 40{.}06 & 40{.}53	& 40{.}64 & \bf{40{.}95}\\
\midrule
\multicolumn{3}{l}{\bf IWSLT15}\\
& Vi-En & 29{.}57  & 29{.}56 & \bf{30{.}01} & \bf{30{.}08}\\
& En-Vi & 31{.}78 & 30{.}99	& \bf{32{.}54} & 31{.}17 \\
\midrule
\multicolumn{6}{l}{\bf WMT14, sacrebleu}\\
&En-De & 26{.}61 & 27{.}01 & 26{.}92 & \bf{27{.}29}\\
&De-En & 30{.}96 & 30{.}15 & \bf{31{.}25} & 30{.}59\\
\midrule
\multicolumn{6}{l}{\bf WMT14, eval as in ~\citep{Attentionallyouneed}}\\
&En-De & 27{.}37 & 27{.}48 & \bf{27{.}71} & \bf{27{.}67}\\
%&De-En & & &  &\\
\bottomrule
\end{tabular}
\caption{BLEU scores. Bold indicates the best score and all scores whose difference from the best is not statistically significant (with $p$-value of 0.05). (Statistical significance is computed via bootstrap~\cite{koehn2004statistical}.)}
\label{tab:main_multisentence_scores}
\end{table}

Table~\ref{tab:main_multisentence_scores} examines the effects of MSR on a range of translation tasks. All scores are computed with Sacrebleu and default beam search from fairseq~\citep{fairseq}, except "WMT$14$, eval as in~\citep{Attentionallyouneed}". In this table, all MSR experiments are conducted with $N\small{=}4$, which is a simple baseline to choose $N$. We can make the following observations. Firstly, on all datasets and translation directions, models trained with \textit{Multi-Sentence Resampling} statistically significantly outperform baselines: from $0.42$ to $0.76$ BLEU on Fr-En, En-Fr, Vi-En and En-Vi, and nearly $0.3$ BLEU for En-De and De-En. Secondly, length-normalized models work significantly better than models without length-normalization only in $2$ directions out of $6$. We did not tune length-normalization hyperparameters in our experiments; however, our results suggest that length-normalization may be unnecessary in some cases.

\subsection{Training time}

As with any regularization, Multi-Sentence Resampling increases the training time of models. Although we expanded the dataset by 10x and 5x times for IWSLT and WMT, respectively, the training time in both cases did not increase with the size of the dataset. Table~\ref{tab:main_multisentence_scores} shows that MSR with $N=4$ increases training time before convergence on average $80\%$ among used datasets, compared to the default training. This suggests that it is possible to make a more memory efficient MSR implementation as part of the data processing pipeline which does MSR on-the-fly, without the need to pre-process a training dataset which is 5-10x larger than the original. However, we leave this to future work. %Figure~\ref{fig:bleu-steps} shows that training with MSR converges slightly slower, but training for longer allows models to converge to better solutions.  
\begin{table}[h!]
\centering
\begin{tabular}{lccc}
\toprule
& & \bf original  & \bf MSR \\
\midrule
\multicolumn{3}{l}{\bf IWSLT17}\\
& Fr-En & 60 & 118  \\
& En-Fr & 77 & 195	\\
\midrule
\multicolumn{3}{l}{\bf IWSLT15}\\
& Vi-En & 55 & 75\\
& En-Vi & 53 & 85\\
\midrule
\multicolumn{3}{l}{\bf WMT14}\\
&En-De & 150 & 320\\
&De-En & 174  & 250\\
\bottomrule
\end{tabular}
\caption{Number of thousands of training steps before convergence for experiments from Table~\ref{tab:main_multisentence_scores}.}
\label{tab:training_steps}
\end{table}

\section{Conclusions}

In this work, we analyzed errors that cause quality degradation with growing beam size in NMT and ASR. We demonstrated that the major contribution to quality degradation on large beams comes from short translations, which are early terminated prefixes of hypotheses which are long when decoding with a small beams. In contrast to ASR, we showed that the reference length on which beam search degradation begins to grow is connected with the low number of sentences longer than this length during training. Thus, usual NMT datasets, that are biased towards short sentences, strongly contribute to the degradation with large beams. Based on this finding, we introduced \textit{Multi-Sentence Resampling} -- a simple data augmentation technique. It concatenates several sentences from a dataset, increasing the length of training examples. Models trained with \textit{Multi-Sentence Resampling} were shown to consistently outperform baseline models on IWSLT$15$ En-Vi, IWSLT$17$ En-Fr, and WMT$14$ En-De datasets. Thus, we demonstrate that it is possible to mitigate beam search degradation with data augmentation. Future research directions include adapting \textit{Multi-Sentence Resampling} to other domains like ASR and studying beam search problems for document-level Machine Translation, where adjacent sentences are naturally connected.

%In this work we analyze the beam-search problem and length bias. We show that most of the quality degradation with growing beam size comes from short translations of long sentences. We also show that these translations are early finalized prefixes of long hypotheses. We connect these problems with the distribution of lengths of sentences in train datasets. Machine Translation train datasets are usually biased towards short sentences, causing models biased toward short translations. To solve this problem, we propose the \textit{Multi-Sentence Resamling} a simple data augmentation method that increases the length of train examples by concatenating multiple train sentences. We show that our method significantly decreases quality degradation with growing beam size and improves the final quality on a range of translation tasks.

\bibliography{bibliography}

\begin{thebibliography}{25}
\expandafter\ifx\csname natexlab\endcsname\relax\def\natexlab#1{#1}\fi

\bibitem[{Bahdanau et~al.(2016)Bahdanau, Cho, and Bengio}]{bahdanau2016neural}
Dzmitry Bahdanau, Kyunghyun Cho, and Yoshua Bengio. 2016.
\newblock \href {http://arxiv.org/abs/1409.0473} {Neural machine translation by
  jointly learning to align and translate}.

\bibitem[{Chorowski and Jaitly(2017)}]{towards-better-decoding-asr}
Jan Chorowski and Navdeep Jaitly. 2017.
\newblock \href {https://arxiv.org/abs/1612.02695} {Towards better decoding and
  language model integration in sequence to sequence models}.

\bibitem[{Ding et~al.(2019)Ding, Renduchintala, and Duh}]{bpe-size-2019}
Shuoyang Ding, Adithya Renduchintala, and Kevin Duh. 2019.
\newblock \href {https://www.aclweb.org/anthology/W19-6620} {A call for prudent
  choice of subword merge operations in neural machine translation}.
\newblock In \emph{Proceedings of Machine Translation Summit XVII Volume 1:
  Research Track}, pages 204--213, Dublin, Ireland. European Association for
  Machine Translation.

\bibitem[{Eikema and Aziz(2020)}]{map-decoding-all-you-need}
Bryan Eikema and Wilker Aziz. 2020.
\newblock \href {http://arxiv.org/abs/2005.10283} {Is map decoding all you
  need? the inadequacy of the mode in neural machine translation}.

\bibitem[{Koehn(2004)}]{koehn2004statistical}
Philipp Koehn. 2004.
\newblock \href {https://www.aclweb.org/anthology/W04-3250} {Statistical
  significance tests for machine translation evaluation}.
\newblock In \emph{Proceedings of the 2004 Conference on Empirical Methods in
  Natural Language Processing}.

\bibitem[{Koehn and Knowles(2017)}]{mt-problems-2017}
Philipp Koehn and Rebecca Knowles. 2017.
\newblock \href {https://doi.org/10.18653/v1/W17-3204} {Six challenges for
  neural machine translation}.
\newblock In \emph{Proceedings of the First Workshop on Neural Machine
  Translation}, pages 28--39, Vancouver. Association for Computational
  Linguistics.

\bibitem[{Kumar and Sarawagi(2019)}]{calibration-of-encoder-decoder}
Aviral Kumar and Sunita Sarawagi. 2019.
\newblock \href {http://arxiv.org/abs/1903.00802} {Calibration of encoder
  decoder models for neural machine translation}.
\newblock \emph{CoRR}, abs/1903.00802.

\bibitem[{Marzal and Vidal(1993)}]{WER-computation}
A.~Marzal and E.~Vidal. 1993.
\newblock \href {https://doi.org/10.1109/34.232078} {Computation of normalized
  edit distance and applications}.
\newblock \emph{IEEE Transactions on Pattern Analysis and Machine
  Intelligence}, 15(9):926--932.

\bibitem[{Meister et~al.(2020)Meister, Vieira, and
  Cotterell}]{beam-search-question}
Clara Meister, Tim Vieira, and Ryan Cotterell. 2020.
\newblock \href {http://arxiv.org/abs/2010.02650} {If beam search is the
  answer, what was the question?}

\bibitem[{Murray and Chiang(2018)}]{correcting-length-bias-2018}
Kenton Murray and David Chiang. 2018.
\newblock \href {https://doi.org/10.18653/v1/W18-6322} {Correcting length bias
  in neural machine translation}.
\newblock In \emph{Proceedings of the Third Conference on Machine Translation:
  Research Papers}, pages 212--223, Brussels, Belgium. Association for
  Computational Linguistics.

\bibitem[{Ott et~al.(2019)Ott, Edunov, Baevski, Fan, Gross, Ng, Grangier, and
  Auli}]{fairseq}
Myle Ott, Sergey Edunov, Alexei Baevski, Angela Fan, Sam Gross, Nathan Ng,
  David Grangier, and Michael Auli. 2019.
\newblock fairseq: A fast, extensible toolkit for sequence modeling.
\newblock In \emph{Proceedings of NAACL-HLT 2019: Demonstrations}.

\bibitem[{Panayotov et~al.(2015)Panayotov, Chen, Povey, and
  Khudanpur}]{librispeech}
Vassil Panayotov, Guoguo Chen, Daniel Povey, and Sanjeev Khudanpur. 2015.
\newblock \href {https://doi.org/10.1109/ICASSP.2015.7178964} {Librispeech: An
  asr corpus based on public domain audio books}.
\newblock In \emph{2015 IEEE International Conference on Acoustics, Speech and
  Signal Processing (ICASSP)}, pages 5206--5210.

\bibitem[{Papineni et~al.(2002)Papineni, Roukos, Ward, and
  Zhu}]{BLEU-papineni-etal-2002}
Kishore Papineni, Salim Roukos, Todd Ward, and Wei-Jing Zhu. 2002.
\newblock \href {https://doi.org/10.3115/1073083.1073135} {{B}leu: a method for
  automatic evaluation of machine translation}.
\newblock In \emph{Proceedings of the 40th Annual Meeting of the Association
  for Computational Linguistics}, pages 311--318, Philadelphia, Pennsylvania,
  USA. Association for Computational Linguistics.

\bibitem[{Post(2018)}]{sacrebleu}
Matt Post. 2018.
\newblock \href {https://www.aclweb.org/anthology/W18-6319} {A call for clarity
  in reporting {BLEU} scores}.
\newblock In \emph{Proceedings of the Third Conference on Machine Translation:
  Research Papers}, pages 186--191, Belgium, Brussels. Association for
  Computational Linguistics.

\bibitem[{Sennrich et~al.(2016)Sennrich, Haddow, and
  Birch}]{sennrich-etal-2016-bpe}
Rico Sennrich, Barry Haddow, and Alexandra Birch. 2016.
\newblock \href {https://doi.org/10.18653/v1/P16-1162} {Neural machine
  translation of rare words with subword units}.
\newblock In \emph{Proceedings of the 54th Annual Meeting of the Association
  for Computational Linguistics (Volume 1: Long Papers)}, pages 1715--1725,
  Berlin, Germany. Association for Computational Linguistics.

\bibitem[{Sountsov and Sarawagi(2016)}]{2016-length-bias}
Pavel Sountsov and Sunita Sarawagi. 2016.
\newblock \href {https://doi.org/10.18653/v1/D16-1158} {Length bias in encoder
  decoder models and a case for global conditioning}.
\newblock In \emph{Proceedings of the 2016 Conference on Empirical Methods in
  Natural Language Processing}, pages 1516--1525, Austin, Texas. Association
  for Computational Linguistics.

\bibitem[{Stahlberg and Byrne(2019)}]{cat-tongue}
Felix Stahlberg and Bill Byrne. 2019.
\newblock \href {https://doi.org/10.18653/v1/D19-1331} {On {NMT} search errors
  and model errors: Cat got your tongue?}
\newblock In \emph{Proceedings of the 2019 Conference on Empirical Methods in
  Natural Language Processing and the 9th International Joint Conference on
  Natural Language Processing (EMNLP-IJCNLP)}, pages 3356--3362, Hong Kong,
  China. Association for Computational Linguistics.

\bibitem[{Vaswani et~al.(2017)Vaswani, Shazeer, Parmar, Uszkoreit, Jones,
  Gomez, Kaiser, and Polosukhin}]{Attentionallyouneed}
Ashish Vaswani, Noam Shazeer, Niki Parmar, Jakob Uszkoreit, Llion Jones,
  Aidan~N Gomez, \L~ukasz Kaiser, and Illia Polosukhin. 2017.
\newblock \href
  {http://papers.nips.cc/paper/7181-attention-is-all-you-need.pdf} {Attention
  is all you need}.
\newblock In I.~Guyon, U.~V. Luxburg, S.~Bengio, H.~Wallach, R.~Fergus,
  S.~Vishwanathan, and R.~Garnett, editors, \emph{Advances in Neural
  Information Processing Systems 30}, pages 5998--6008. Curran Associates, Inc.

\bibitem[{Wang et~al.(2020{\natexlab{a}})Wang, Tang, Ma, Wu, Okhonko, and
  Pino}]{fairseq-s2t}
Changhan Wang, Yun Tang, Xutai Ma, Anne Wu, Dmytro Okhonko, and Juan Pino.
  2020{\natexlab{a}}.
\newblock \href {https://www.aclweb.org/anthology/2020.aacl-demo.6} {Fairseq
  {S}2{T}: Fast speech-to-text modeling with fairseq}.
\newblock In \emph{Proceedings of the 1st Conference of the Asia-Pacific
  Chapter of the Association for Computational Linguistics and the 10th
  International Joint Conference on Natural Language Processing: System
  Demonstrations}, pages 33--39, Suzhou, China. Association for Computational
  Linguistics.

\bibitem[{Wang and Sennrich(2020)}]{wang-sennrich-exposure-2020}
Chaojun Wang and Rico Sennrich. 2020.
\newblock \href {https://doi.org/10.18653/v1/2020.acl-main.326} {On exposure
  bias, hallucination and domain shift in neural machine translation}.
\newblock In \emph{Proceedings of the 58th Annual Meeting of the Association
  for Computational Linguistics}, pages 3544--3552, Online. Association for
  Computational Linguistics.

\bibitem[{Wang et~al.(2020{\natexlab{b}})Wang, Tu, Shi, and
  Liu}]{inference-calib-2020}
Shuo Wang, Zhaopeng Tu, Shuming Shi, and Yang Liu. 2020{\natexlab{b}}.
\newblock \href {https://doi.org/10.18653/v1/2020.acl-main.278} {On the
  inference calibration of neural machine translation}.
\newblock In \emph{Proceedings of the 58th Annual Meeting of the Association
  for Computational Linguistics}, pages 3070--3079, Online. Association for
  Computational Linguistics.

\bibitem[{Wu et~al.(2016)Wu, Schuster, Chen, Le, Norouzi, Macherey, Krikun,
  Cao, Gao, Macherey, Klingner, Shah, Johnson, Liu, Kaiser, Gouws, Kato, Kudo,
  Kazawa, Stevens, Kurian, Patil, Wang, Young, Smith, Riesa, Rudnick, Vinyals,
  Corrado, Hughes, and Dean}]{Wu2016GooglesNM}
Y.~Wu, Mike Schuster, Z.~Chen, Quoc~V. Le, Mohammad Norouzi, Wolfgang Macherey,
  M.~Krikun, Yuan Cao, Q.~Gao, Klaus Macherey, Jeff Klingner, Apurva Shah,
  M.~Johnson, X.~Liu, L.~Kaiser, S.~Gouws, Y.~Kato, Taku Kudo, H.~Kazawa,
  K.~Stevens, G.~Kurian, Nishant Patil, W.~Wang, C.~Young, J.~Smith, Jason
  Riesa, Alex Rudnick, Oriol Vinyals, G.~S. Corrado, Macduff Hughes, and
  J.~Dean. 2016.
\newblock Google's neural machine translation system: Bridging the gap between
  human and machine translation.
\newblock \emph{ArXiv}, abs/1609.08144.

\bibitem[{Yang et~al.(2018)Yang, Huang, and Ma}]{breaking-beam-curse-2019}
Yilin Yang, Liang Huang, and Mingbo Ma. 2018.
\newblock \href {https://doi.org/10.18653/v1/D18-1342} {Breaking the beam
  search curse: A study of (re-)scoring methods and stopping criteria for
  neural machine translation}.
\newblock In \emph{Proceedings of the 2018 Conference on Empirical Methods in
  Natural Language Processing}, pages 3054--3059, Brussels, Belgium.
  Association for Computational Linguistics.

\bibitem[{Yang et~al.(2020)Yang, Gao, Wang, and Ney}]{predicting-length-2020}
Zijian Yang, Yingbo Gao, Weiyue Wang, and Hermann Ney. 2020.
\newblock \href {https://www.aclweb.org/anthology/2020.aacl-main.41}
  {Predicting and using target length in neural machine translation}.
\newblock In \emph{Proceedings of the 1st Conference of the Asia-Pacific
  Chapter of the Association for Computational Linguistics and the 10th
  International Joint Conference on Natural Language Processing}, pages
  389--395, Suzhou, China. Association for Computational Linguistics.

\bibitem[{Zhou et~al.(2020)Zhou, Schlüter, and
  Ney}]{renormalize-prob-pruning-beam-asr}
Wei Zhou, Ralf Schlüter, and Hermann Ney. 2020.
\newblock Robust beam search for encoder-decoder attention based speech
  recognition without length bias.
\newblock In \emph{Interspeech}, pages 1768--1772, Shanghai, China.

\end{thebibliography}
\bibliographystyle{acl_natbib}

\newpage
\appendix
\label{sec:appendix}

\end{document}